\documentclass[sigconf]{acmart}
\AtBeginDocument{%
  \providecommand\BibTeX{{%
    \normalfont B\kern-0.5em{\scshape i\kern-0.25em b}\kern-0.8em\TeX}}}

\usepackage{microtype}

\usepackage{graphicx}
\usepackage{amsmath}
\usepackage{amssymb}
\usepackage{amsthm}
\usepackage{booktabs}
\usepackage{algorithm}
\usepackage{algorithmic}
\usepackage{dsfont}
\usepackage{multirow}
\usepackage{bm}
\usepackage{pifont}
\usepackage{bbding}
\newcommand{\tabincell}[2]{\begin{tabular}{@{}#1@{}}#2\end{tabular}}

\newcommand\blfootnote[1]{%
\begingroup 
\renewcommand\thefootnote{}\footnote{#1}%
\addtocounter{footnote}{-1}%
\endgroup 
}

\def\ie{{\it i.e.}}

\setcopyright{acmcopyright}

\copyrightyear{2022}
\acmYear{2022}
\setcopyright{acmcopyright}\acmConference[MM '22]{Proceedings of the 30th ACM
International Conference on Multimedia}{October 10--14, 2022}{Lisboa, Portugal}
\acmBooktitle{Proceedings of the 30th ACM International Conference on Multimedia
(MM '22), October 10--14, 2022, Lisboa, Portugal}
\acmPrice{15.00}
\acmDOI{10.1145/3503161.3547782}
\acmISBN{978-1-4503-9203-7/22/10}

\begin{document}
\title{Skimming, Locating, then Perusing: A Human-Like Framework for Natural Language Video Localization}

\author{Daizong Liu}
\email{dzliu@stu.pku.edu.cn}
\affiliation{%
  \institution{Wangxuan Institute of Computer \\
Technology, Peking University}
  \city{Beijing}
  \country{China}
  }

\author{Wei Hu$^\dagger$}
\email{forhuwei@pku.edu.cn}
\affiliation{%
  \institution{Wangxuan Institute of Computer \\
Technology, Peking University}
  \city{Beijing}
  \country{China}
  }

\renewcommand{\shortauthors}{Daizong Liu \& Wei Hu}

\begin{abstract}
This paper addresses the problem of natural language video localization (NLVL).
Almost all existing works follow the ``only look once" framework that exploits a single model to directly capture the complex cross- and self-modal relations among video-query pairs and retrieve the relevant segment. However, we argue that these methods have overlooked two indispensable characteristics of an ideal localization method:
1) Frame-differentiable: considering the imbalance of positive/negative video frames, it is effective to highlight positive frames and weaken negative ones during the localization.
2) Boundary-precise: to predict the exact segment boundary, the model should capture more fine-grained differences between consecutive frames since their variations are often smooth.
To this end, inspired by how humans perceive and localize a segment, we propose a two-step human-like framework called Skimming-Locating-Perusing (SLP). 
SLP consists of a Skimming-and-Locating (SL) module and a Bi-directional Perusing (BP) module. 
The SL module first refers to the query semantic and selects the best matched frame from the video while filtering out irrelevant frames. Then, the BP module constructs an initial segment based on this frame, and dynamically updates it by exploring its adjacent frames until no frame shares the same activity semantic. 
Experimental results on three challenging benchmarks show that our SLP is superior to the state-of-the-art methods and localizes more precise segment boundaries.
\end{abstract}



\maketitle

\blfootnote{
$^\dagger$This work is supported by the National Key R\&D Program of China under contract No. 2021YFF0901502.
Corresponding author: Wei Hu (forhuwei@pku.edu.cn).}

\section{Introduction}
\begin{figure}[t!]
\centering
\includegraphics[width=0.50\textwidth]{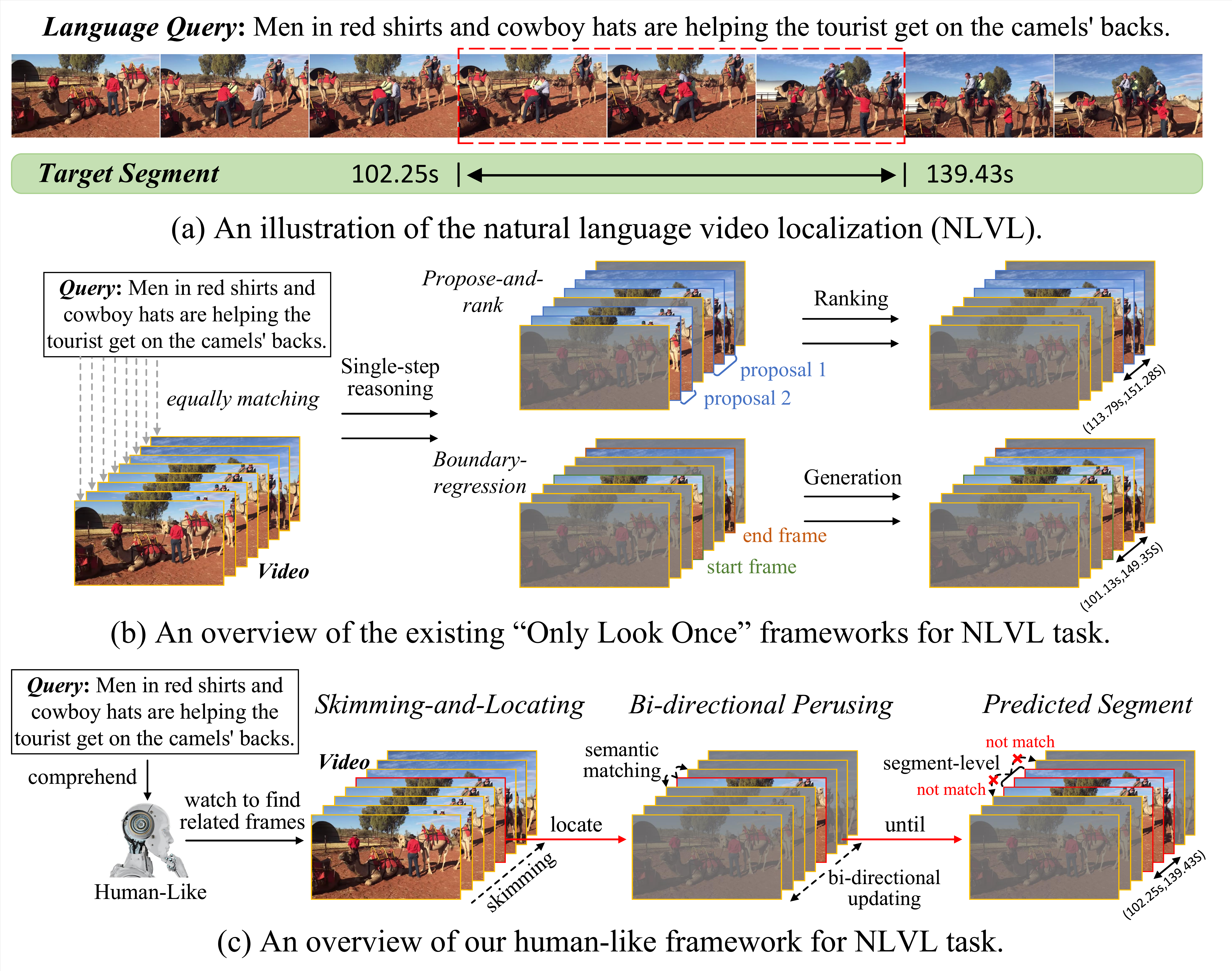}
\vspace{-16pt}
\caption{(a) An illustrative example of the NLVL task. (b) The overview of the existing ``only look once" frameworks, which equally learn the frame-word and frame-frame relations and predict the segment via a single-step reasoning model. (c) The overview of our human-like framework, which first skims the entire video to locate the most relevant frame according to the query semantic, then peruses its adjacent frames to construct and update more accurate segment.}
\label{fig:introduction}
\vspace{-10pt}
\end{figure}

Natural language video localization (NLVL) is an important yet challenging task in multimedia understanding, which has drawn increasing attention in recent years due to its vast potential applications in information retrieval \cite{yadav2020unsupervised,mao2021generation} and human-computer interaction \cite{singha2018dynamic,li2021gwlan}. 
This task aims to localize a specific video segment that semantically corresponds to a sentence query, as shown in Figure~\ref{fig:introduction} (a). 
NLVL requires not only the modeling of the complex multi-modal interactions among vision and language, but also the characterization of complicated contexts for semantic alignment.

Most existing works \cite{anne2017localizing,gao2017tall,zhang2019cross,yuan2019semantic,zhang2019learning} tackle the NLVL following a \textit{propose-and-rank} architecture, which first predefines multiple segment candidates of different time intervals and then ranks them according to their semantic similarities with the query. The best segment is selected based on the similarities. These methods are sensitive to the quality of segment candidates and are inefficient due to redundant candidate-query matching. Recently, several works \cite{chenrethinking,yuan2019find,zhang2020span,zhang2021parallel} exploit a more efficient \textit{boundary-regression} architecture that directly predicts the start/end boundaries of the target segment.
However, they neglect the rich internal information between start and end boundaries without capturing the segment-level interaction.
Whether adopting \textit{propose-and-rank} or \textit{boundary-regression} architecture, almost all current methods follow the ``only look once" strategy that equally learns the relations between all frame-word and frame-frame pairs of the matched video-query within a single modeling process. The segment is subsequently predicted based on all query-guided frame features. Details are shown in Figure~\ref{fig:introduction} (b), 

In this paper, we argue that these methods are unnatural to human perceptions and have overlooked two indispensable characteristics of an ideal localization strategy:
1) Frame-differentiable: A video often contains thousands of frames, but maybe only a few positive frames are related to the query. Instead of predicting the segments by equally considering all frames, it is more effective to highlight positive frames and weaken negative ones during the localization.
2) Boundary-precise: It is challenging to distinguish visually similar contents in consecutive frames of a video with the ``only look once" strategy, since the variation between the adjacent frames is often smooth.
Instead of relying on the single-step reasoning,
it is required to deeply mine more fine-grained differences between adjacent frames for precisely locating the segment boundaries.
Luckily, the above two characteristics have already been captured by human brain and are the key reasons why humans are able to retrieve the exact segment effectively. 
As shown in Figure~\ref{fig:introduction} (c), given a video-query pair, humans generally first comprehend the query semantic and skim through the video to locate the most matched frame. Based on this matched frame, humans then gradually peruse and decide whether its adjacent frames share the same semantic related to the target activity. The predicted segment is determined when no adjacent frames match the activity semantic. Such human-aware localization strategy can also alleviate the problems of redundant candidates and lacking internal information in \textit{propose-and-rank} and \textit{boundary-regression} frameworks.

To this end, inspired by how humans localize a segment in a video, we propose a human-like framework for NLVL---Skimming-Locating-Perusing (SLP), 
which develops a ``two-step" localization strategy to determine accurate segment boundaries. 
We first develop a Skimming-and-Locating (SL) module to imitate human perceptions that locate the most likely positive frames corresponding to query semantic by skimming through the entire video. 
In SL module, we comprehend the aligned semantics between the video-query pair and highlight the relevant frames as well as suppress irrelevant ones for discriminative frame-wise representation learning, which is realized by a query-guided content comprehension network. 
Then, we propose a Bi-directional Perusing (BP) module to dynamically construct segments based on the previously predicted positive frames. Specifically, the BP module takes each predicted positive frame as the initial segment, and then gradually adds the adjacent frames into the segment if they share the same linguistic query or visual appearance semantic. This perusing strategy exploits fine-grained differences among consecutive frames, leading to more precise segment boundary localization.

Our main contributions are summarized as follows:
\begin{itemize}
    \item Firstly, we propose a novel Skimming-Locating-Perusing (SLP), which is the first human-like framework for NLVL to take both frame-differentiable and boundary-precise requirements into account to the best of our knowledge.
    \item Secondly, different from the ``only look once" localization strategy, our ``two-step” SL and BP modules highlight more impact on the positive frames and capture more fine-grained differences between adjacent frames. 
    \item Finally, experiments validate the effectiveness of SLP over three challenging benchmarks ActivityNet Caption, TACoS and Charades-STA.
\end{itemize}

\section{Related Works}
\noindent \textbf{Natural language image retrieval.} Early works of localization task mainly focus on localizing the image region corresponding to a language query. They first generate candidate image regions using image proposal method \cite{ren2015faster}, and then find the matched one with respect to the given query. Some works \cite{mao2016generation,hu2016natural,rohrbach2016grounding} try to extract target image regions based on description reconstruction error or probabilities. There are also several studies \cite{yu2016modeling,chen2017query,chen2017msrc,zhang2018grounding} considering incorporating contextual information of region-phrase relationship into the localization model. \cite{wang2016structured} further models region-region and phrase-phrase structures. Some other methods exploit attention modeling in queries, images, or object proposals \cite{endo2017attention,deng2018visual,yu2018mattnet}.

\noindent \textbf{Natural language video localization.}
Natural language video localization (NLVL) is a new task introduced recently \cite{gao2017tall,anne2017localizing}.
Various algorithms \cite{anne2017localizing,gao2017tall,chen2018temporally,zhang2019cross,yuan2019semantic,zhang2019learning,liu2021context,liu2021progressively,liu2020jointly,liu2020reasoning,liu2021adaptive,liu2022memory,liu2022exploring} have been proposed within the \textit{propose-and-rank} framework, which first generates segment candidates and then utilizes multimodal matching to retrieve the most relevant candidate for a query. Some of them \cite{anne2017localizing,gao2017tall} take multiple sliding windows as candidates. To improve the quality of the candidates, \cite{zhang2019cross,yuan2019semantic} pre-cut the video on each frame by multiple pre-defined temporal scales, and directly integrate sentence information with fine-grained video clip for scoring.
For instance, Xu \textit{et al.} \cite{xu2019multilevel} introduce a multi-level model to integrate visual and textual features earlier and further re-generate queries as an auxiliary task.
Chen \textit{et al.} \cite{chen2018temporally} capture the evolving fine-grained frame-by-word interactions between video and query to enhance the video representation understanding.
Zhang \textit{et al.} \cite{zhang2019man} model relations among candidate segments produced from a convolutional neural network with the guidance of the query information.
Although these methods achieve great performance, they are severely limited by the heavy computation on proposal matching/ranking, and sensitive to the quality of pre-defined proposals.

Recently, many methods \cite{rodriguez2020proposal,chenrethinking,yuan2019find,mun2020LGI,zeng2020dense,zhang2020span,nan2021interventional,zhang2021parallel,liu2022unsupervised,liu2022reducing} propose to utilize the \textit{boundary-regression} framework. Specifically, instead of relying on the segment candidates, they directly predict two probabilities at each frame by leveraging cross-modal interactions between video and query, which indicate whether this frame is a start/end frame of the ground truth video segment.
There are also some reinforcement learning (RL) based frameworks \cite{hahn2019tripping,wu2020tree} proposed in the NLVL task.

\begin{figure*}[t!]
\centering
\includegraphics[width=1.0\textwidth]{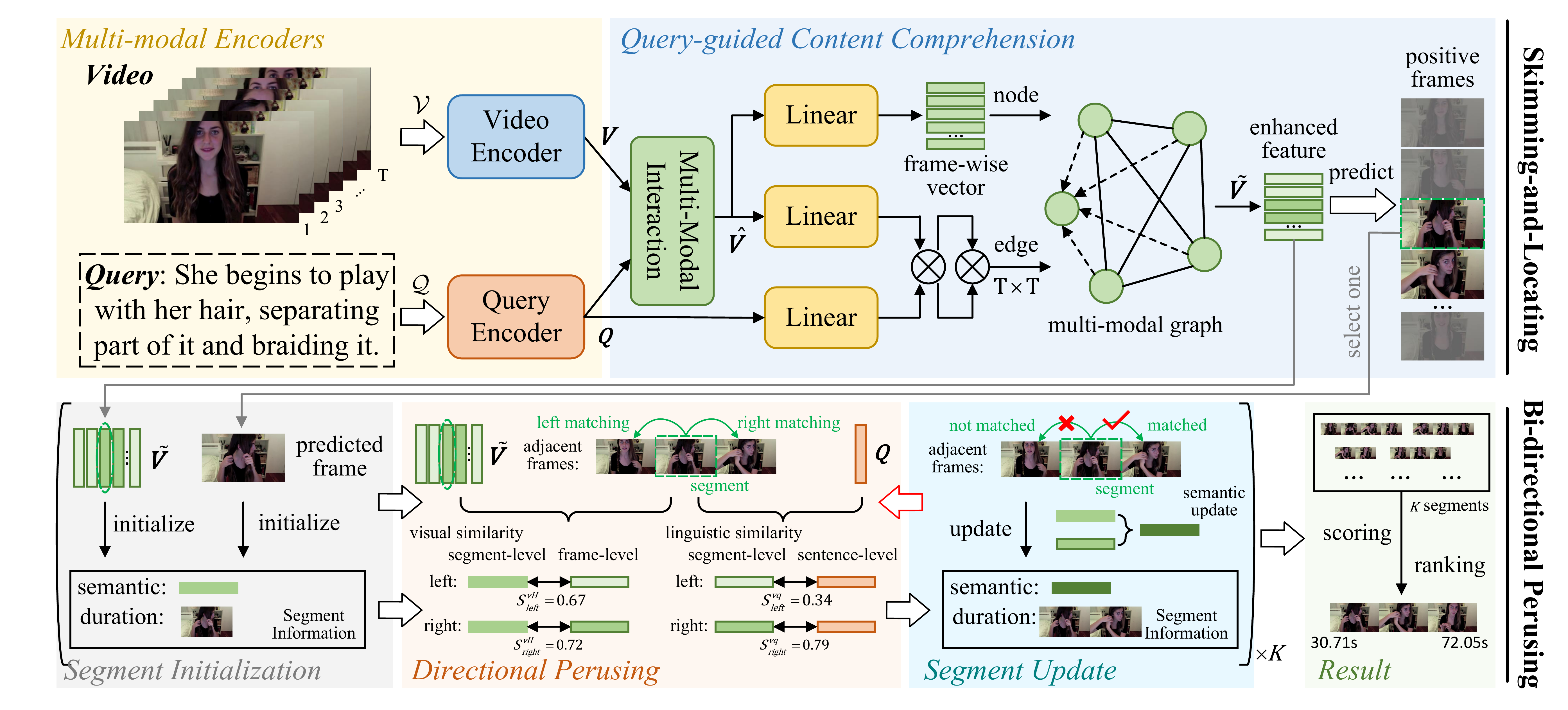}
\caption{An overview of the proposed Skimming-Locating-Perusing (SLP) architecture for NLVL. The Skimming-and-Locating (SL) module first predicts the top-$K$ ranked positive frames among the entire video. Then, the Bi-directional Perusing (BP) module dynamically constructs and updates the segment based on each predicted positive frame. The final segment is selected based on the learned confidence score. }
\vspace{-10pt}
\label{fig:pipeline}
\end{figure*}

However, almost all the above two classes of methods equally learn the complex frame-to-word and frame-to-frame relations within a single modeling process and retrieve the segment based on all frames, which often fails to distinguish the foreground-background frames and capture the fine-grained differences between certain consecutive frames for determining precise segment boundaries.

\section{Methodology}
\subsection{Problem Statement}
Given an untrimmed video $\mathcal{V}$, we represent it as $\mathcal{V}=\{v_t\}_{t=1}^T$  frame-by-fram, where $v_t$ is the $t$-th frame and $T$ is the length of the entire video. We denote the given sentence query with $N$ words as $\mathcal{Q}=\{q_n\}_{n=1}^N$ word-by-word, where $q_n$ is the $n$-th word in the query. The NLVL task aims to localize a segment in video $\mathcal{V}$ starting at timestamp $\tau_s$ and ending at timestamp $\tau_e$, which corresponds to the query $\mathcal{Q}$ semantically. This task is quite challenging since it is difficult to localize activities of interest precisely in a long video with complex contents.

\subsection{Overview}
We focus on addressing the issue that existing NLVL methods with the ``only look once" strategy are unnatural to human perceptions and often fail to mine more fine-grained differences between adjacent frames for precise boundary prediction. 
To this end, we propose a human-like framework to imitate how humans localize the segment associated to the query. 
Instead of following the ``only look once” strategy to utilize a single-step reasoning process, we introduce a novel two-step localization paradigm of ``Skimming-Locating-Perusing" (SLP), which consists of two main modules as shown in Figure~\ref{fig:pipeline}:
(1) {\bf Skimming-and-Locating (SL) module:} It first predicts the most query-related positive frames by fine-grained multi-modal reasoning for filtering out negative ones;
(2) {\bf Bi-directional Perusing (BP) module:} After obtaining the positive frames, it then dynamically constructs desired segment by finely searching and adding semantically similar adjacent frames based on each predicted positive frame.
In such a coarse-to-fine manner, we are able to determine more precise segment boundaries.

Formally, we first forward the input video $\mathcal{V}=\{v_t\}_{t=1}^T$ and the query $\mathcal{Q}=\{q_n\}_{n=1}^N$ to the SL module for multi-modal feature encoding and interaction. 
Then it selects the foremost matched frames related to the query as the positive frames by:
\begin{equation}
    \{y_t\}_{t=1}^{T} = \text{SL}(\mathcal{V},\mathcal{Q}),
\end{equation}
where $y_t \in \{0,1\}$ is a binary class to determine whether the $t$-th frame is positive or negative. 

With the predicted frame-wise class $\{y_t\}_{t=1}^{T}$, the BP module constructs the desired segment based on each predicted positive frame by bi-directionally searching its adjacent frames sharing the same activity semantic. 
In particular, it first takes one of the predicted positive frames as the initial segment and then dynamically updates this segment by:
\begin{equation}
\begin{aligned}
   & \quad \quad \quad (t-1,t+1) = \text{BP}((t,t), \mathcal{V}, \mathcal{Q}), \\
   & \quad \quad \quad \quad \quad \quad \quad \quad ...... \\
   & (t-l_s,t+l_e) = \text{BP}((t-l_s,t+l_e-1), \mathcal{V}, \mathcal{Q}),
\end{aligned}
\end{equation}
where $t$ is the index of the selected positive frame and $(t,t)$ denotes the initial segment. The final predicted segment $(t-l_s,t+l_e)$ is constructed by iteratively searching the adjacent frames near the segment boundary and updating the segment with the semantically related adjacent frames.
Specifically, whether adjacent frames near the boundary are added to the segment depends on whether they share both the same query-guided and visual-related semantics to the segment.

\noindent \textbf{Inference details.}
Given an input video-query pair, we first forward them through the SL module to predict top $K$ positive frames. Then, for each positive frame, we initialize the corresponding segment and deploy the BP module to dynamically update it. The final localization result is obtained by choosing the segment among the $K$ constructed ones with the highest confidence score.

In the following, we first elaborate on the SL module and BP module, respectively. Then, we present the training details of the proposed SLP model.

\subsection{Skimming-and-Locating Module}
\noindent \textbf{Multi-modal encoders.} Given a video $\mathcal{V}=\{v_t\}_{t=1}^T$ with $T$ frames, we first extract its features by a pre-trained C3D
network \cite{tran2015learning}, and then employ a multi-head self-attention \cite{vaswani2017attention} module to capture the long-range dependencies among the video frames. We also utilize a Bi-GRU \cite{chung2014empirical} layer to learn the sequential characteristic. The final video features are denoted as $\bm{V}=\{\bm{v}_t\}_{t=1}^T \in \mathbb{R}^{T\times D}$, where $D$ is the feature dimension. 
For query $\mathcal{Q}$ encoding, following previous works \cite{zhang2019cross,zeng2020dense}, we first generate the word-level embeddings using the Glove \cite{pennington2014glove} embeddings, and then employ another multi-head self-attention module and Bi-GRU layer to further encode the query features as $\bm{Q}=\{\bm{q}_n\}_{n=1}^N \in \mathbb{R}^{N\times D}$.

\noindent \textbf{Query-guided content comprehension.}
To skim and locate the most positive frames related to the query semantic, it is natural to learn and enhance the video-query interaction for frame-wise discriminative representation learning. Therefore, we propose a query-guided content comprehension (QCC) submodule which consists of two stages: 1) We first associate frame-wise video features with the correlated word-wise query features to obtain the multi-modal features. 2) Then, we construct a multi-modal graph over the multi-modal features by connecting frame-wise nodes referring the query-guided semantic similarity of neighbors. 
By reasoning over this graph, the responses of the positive frames matched with the query cue are highlighted while those of non-matched ones are suppressed accordingly. Finally, the enhanced multi-modal features are utilized to predict the positive frames. Comparing with previous works equally learns the frame-wise relations, our QCC module provides more fine-grained frame-wise semantic understanding by query-relevant/irrelevant frame distinguishment for predicting more accurata positive frames.

In the first stage, given the encoded features of video and query $\bm{V},\bm{Q}$, we adopt a typical attention mechanism to associate query information with the video features of each frame. Concretely, we first compute the attention score between each pair of frame and word, and obtain a video-to-query attention matrix $\bm{M} = \{m_{t,n}\}_{t=1,n=1}^{t=T,n=N} \in \mathbb{R}^{T\times N}$. Each score $m_{t,n}$ represents the correlation of the $t$-th frame and $n$-th word, and is formulated by:
\begin{equation}
    m_{t,n} = \bm{w}^{\top} \text{tanh}(\bm{W}_1^m \bm{v}_t + \bm{W}_2^m \bm{q}_n + \bm{b}^m),
\end{equation}
where $\bm{W}_1^m,\bm{W}_2^m,\bm{b}^m$ are learnable parameters, and $\bm{w}^{\top}$ is a row vector \cite{zhang2019cross}. The query-guided frame features $\widehat{\bm{v}}_t$ is obtained by:
\begin{equation}
    \widehat{\bm{v}}_t = [\bm{v}_t;\sum_{n=1}^N \text{softmax}(m_{t,n}) (\bm{W}^Q\bm{q}_n)],
\end{equation}
where $\bm{W}^Q$ projects query contexts into the video domain.
By integrating both video and query contexts into the multi-modal features $\widehat{\bm{V}}=\{\widehat{\bm{v}}_t\}_{t=1}^T \in \mathbb{R}^{T \times 2D}$, all foreground frames that might be referred to the query semantic are perceived appropriately.

In the second stage, we regard each query-guided frame feature $\widehat{\bm{v}}_t$ as a node and construct a fully-connected graph which is composed of $T$ nodes. In order to selectively highlight the positive frames while weakening the background ones, we define the edge weights depending on affinities between connected frames and query semantic. 
The adjacency matrix $\bm{A} \in \mathbb{R}^{T\times T}$ is formulated as:
\begin{equation}
\begin{aligned}
    & \quad \bm{B} = (\widehat{\bm{V}} \bm{W}^A_1) (\widehat{\bm{Q}} \bm{W}^A_2)^{\top},  \\
    & \bm{A} = \text{softmax}(\bm{B}) \text{softmax}(\bm{B}^{\top}),
\end{aligned}
\end{equation}
where $\bm{W}^A_1,\bm{W}^A_2$ are learnable parameters, and $\bm{B} \in \mathbb{R}^{T\times N}$. Each element of $\bm{A}$ represents the normalized magnitude of information flow from one frame to another, which depends on their affinities with the query semantic. Therefore, we apply a graph convolution layer \cite{kipf2016semi} to enhance discriminative representation learning:
\begin{equation}
    \widetilde{\bm{V}} = (\bm{A}+\bm{I}) \widehat{\bm{V}} \bm{W}^A_3,
\end{equation}
where $\bm{I}$ is an identity matrix for adding self-loops. $\widetilde{\bm{V}} \in \mathbb{R}^{T\times D}$ is the enhanced features, achieving the ``skimming" process analogous to the comprehension of query-guided video contents by humans.

Subsequently, we apply a binary classification function on the obtained multi-modal features $\widetilde{\bm{V}}$ to ``locate" the most probable positive video frames falling into the ground-truth segment. 
To this end, we design a binary classification module with three linear layers to predict the class $y_t$ on each frame $t$. Since we capture the fine-grained query-guided frame-to-frame relations in the ``skimming" process, the learned multi-modal feature $\widetilde{\bm{V}}$ is able to predict the positive frames well (demonstrated by the visualization results in the experiments). We define the binary cross-entropy loss as:
\begin{equation}
\label{eq:loss1}
\small
    \mathcal{L}_{class} = - \frac{1}{T} \sum_{t=1}^{T}y_t^{gt}\text{log}(y_t)+ (1-y_t^{gt})\text{log}(1-y_t),
\end{equation}
where $y_t^{gt}$ is the ground truth. 
The predicted top-$K$ ranked frames are taken as the positive frames.

\subsection{Bi-directional Perusing}
\noindent \textbf{Segment initialization.}
Given one of the top-$K$ ranked frames indexed at time $t$ predicted by the features $\widetilde{\bm{V}}$, we take it as the initial segment $(t,t)$ and initialize its feature as $\bm{H}_{t:t}=\widetilde{\bm{v}}_t \in \mathbb{R}^{D}$, where $t:t$ denotes the segment duration. The semantic feature of the segment $\bm{H}_{t:t}$ will be updated by dynamically adding semantically relevant adjacent frames to the it.

\noindent \textbf{Directional perusing.}
Based on the initial segment $\bm{H}_{t:t}$, there are three human-like ways to dynamically peruse the adjacent frames of the current segment: 1) perusing the left frames until no frame matches the target semantic before perusing the right ones; 2) perusing the right frames until no frame matches the target semantic before perusing the left ones; 3) perusing the left and right frames at the same time. Here, we take the first perusing way as an example to illustrate our directional perusing process. Analysis of adopting other ways will be shown in our experiments.

Given the current segment $\bm{H}_{t:t}$ and its left adjacent frame $\widetilde{\bm{v}}_{t-1}$, we formulate a matching score $S(\widetilde{\bm{v}}_{t-1},\bm{H}_{t:t})$ to determine whether the $(t-1)$-th frame should be added into the current segment. We define this matching score $S(\widetilde{\bm{v}}_{t-1},\bm{H}_{t:t})$ as:
\begin{equation}
\label{eq:score}
    S(\widetilde{\bm{v}}_{t-1},\bm{H}_{t:t}) = \alpha_1 S^{vq}_{t-1,t:t} + \alpha_2 S^{vH}_{t-1,t:t},
\end{equation}
where $\alpha_1,\alpha_2$ are balancing parameters. The matching score is composed of two terms: 1) $S^{vq}_{t-1,t:t}$ measures the {\it linguistic similarity} between the semantics of frame $\widetilde{\bm{v}}_{t-1}$ and the query $\bm{Q}$; 2) $S^{vH}_{t-1,t:t}$ measures the {\it visual similarity} between the semantics of frame $\widetilde{\bm{v}}_{t-1}$ and segment $\bm{H}_{t:t}$. 
Both two semantic similarities are crucial to distinguish the query-relevant/irrelevant adjacent frames.
In particular, we employ cosine similarity to formulate them as:
\begin{equation}
\begin{aligned}
    & S^{vq}_{t-1,t:t} = \sum_{n=1}^N  \cos{(\widetilde{\bm{v}}_{t-1},\bm{q}_n)} / N, \\
    & \quad S^{vH}_{t-1,t:t} = \cos{(\widetilde{\bm{v}}_{t-1},\bm{H}_{t:t})}.
\end{aligned}
\end{equation}

To supervise both above frame-query and frame-segment matching during the training, we adopt a hinge-based triplet ranking loss \cite{karpathy2015deep} to encourage the similarity score of matched pairs to be larger than those of mismatched pairs as follows:
\begin{equation}
\label{eq:loss2}
\begin{aligned}
    & \ \ \ \mathcal{L}_{match}^{vq} = \text{max} (0,\beta_1-S^{vq}+S^{\bar{v}q}), \\
    & \quad \quad \quad \quad \ \ + \text{max} (0,\beta_1-S^{vq}+S^{v\bar{q}}), \\
    & \ \ \mathcal{L}_{match}^{vH} = \text{max} (0,\beta_2-S^{vH}+S^{\bar{v}H}) \\
    & \quad \quad \quad \quad \ \ + \text{max} (0,\beta_2-S^{vH}+S^{v\bar{H}}), \\
    & \mathcal{L}_{match} = \text{avg} (\sum \gamma_1 \mathcal{L}_{match}^{vq} + \gamma_2 \mathcal{L}_{match}^{vH}),
\end{aligned}
\end{equation}
where $\beta_1,\beta_2$ and $\gamma_1,\gamma_2$ are the margin and balancing parameters, respectively. $S^{vq}$ is the score of the matched frame-query, $S^{\bar{v}q},S^{v\bar{q}}$ are the scores of the mismatched frame-query. Analogously, $S^{vH}$ is the score of the matched frame-segment, while $S^{\bar{v}H},S^{v\bar{H}}$ are the mismatched ones. Note that, we define the random negative frame/segment (outside the ground-truth segment) as the mismatched frame or mismatched segment, and denote all segment candidates within the ground-truth segment as the matched segments. 
$\text{avg}(\cdot)$ is the average operation.

\noindent \textbf{Segment update.}
Once the above matching score $S(\widetilde{\bm{v}}_{t-1},\bm{H}_{t:t})$ is larger than a pre-defined threshold $\Theta$, we will add the adjacent frame $\widetilde{\bm{v}}_{t-1}$ into the segment, leading to an expanded segment $\bm{H}_{t-1:t} \in \mathbb{R}^{D}$.
Instead of directly adopting max-pooling to update the segment as in \cite{zhang2019learning}, we introduce a new and learnable segment updating strategy to keep the most semantic-related discriminative information and filter out the unimportant one for each new frame since the target activity mostly happens in a local region of the whole frame (the other regions provide redundant information).

Specifically, we first compute the intermediate segment semantic by adding the content of new frame:
\begin{equation}
\label{eq:update1}
\small
    (\bm{H}_{t-1:t})' = \text{tanh}(\bm{W}^H_1(\bm{r}_1 \odot \widetilde{\bm{v}}_{t-1})+\bm{U}^H_1(\bm{r}_2 \odot \bm{H}_{t:t})+\bm{b}^H_1),
\end{equation}
where $\bm{W}_1^H,\bm{U}_1^H$ and $\bm{b}_1^H$ are weights and bias, $\odot$ is an element-wise multiplication. $\bm{r}_1,\bm{r}_2$ are the reset gates learned to forget the background contents in each frame, which are formulated as:
\begin{equation}
\label{eq:update2}
    \bm{r}_{i} = \text{sigmoid}(\bm{W}^r_{i}\widetilde{\bm{v}}_{t-1}+ \bm{U}^r_{i}\bm{H}_{t:t}+\bm{b}^r_{i}), i=1,2
\end{equation}
where $\bm{W}^r_{i},\bm{U}^r_{i}$ and $\bm{b}^r_{i}$ are weights and bias. To decide how much the intermediate semantic $(\bm{H}_{t-1:t})'$ updates the previous segment $\bm{H}_{t:t}$, we develop a gate $\bm{z}$ similarly to the reset gate $\bm{r}_{i}$ and formulate the updating process as:
\begin{equation}
\label{eq:update3}
\begin{aligned}
    \bm{z} = \text{sigmoid}(\bm{W}^z\widetilde{\bm{v}}_{t-1}+ \bm{U}^z\bm{H}_{t:t}+\bm{b}^z), \\
    \bm{H}_{t-1:t} = \bm{z} \odot (\bm{H}_{t-1:t})' + (1-\bm{z}) \odot \bm{H}_{t:t},
\end{aligned}
\end{equation}
where $\bm{W}^z,\bm{U}^z$ and $\bm{b}^z$ are weights and bias. 
The updated segment $\bm{H}_{t-1:t}$ focuses on more precise activity-guided semantics related to the query.
We iteratively update segment using the BP module until no adjacent frames match the activity semantic.

To supervise the segment updating and indicate the quality of each construct segment, we introduce an IoU regression head based on each segment feature $\bm{H}_{t-1:t}$ to predict its confidence score. Specifically, we train a three-layer linear network and acquire the corresponding confidence score $c_{t-1:t}$. The training target $c_{t-1:t}^{gt}$ is obtained by calculating the IoU between segment $(t-1,t)$ and the ground-truth segment $(\tau_s,\tau_e)$. We define the regression loss function as:
\begin{equation}
\label{eq:loss3}
    \mathcal{L}_{conf} = \text{avg}(\sum \mathcal{R}_1(c_{t-1:t},c_{t-1:t}^{gt})),
\end{equation}
where $\mathcal{R}_1$ is the Smooth-L1 loss \cite{girshick2015fast}.
Therefore, we can predict the confidence scores of the final $K$ constructed segment and choose the best one as the result according to their scores.

\subsection{Training Details}
To ensure both SL and BP modules properly, we develop a three-stage training algorithm for our proposed SLP model. 
\textbf{In stage 1}, since the positive frame localization process (binary classification) in SL module often fails to produce high-quality predictions at the beginning of training, we leave the BP module out of this stage and only train the SL module by minimizing Equation (\ref{eq:loss1}) for warming up.
\textbf{In stage 2}, we first fix the SL module obtained in stage 1, and then only train the BP module. Specifically, we enumerate possible frame-query and frame-segment pairs to minimize Equation (\ref{eq:loss2}), and update possible segments initialized from random positive frames for minimizing Equation (\ref{eq:loss3}).
\textbf{In stage 3}, we fine-tune the entire model in an end-to-end manner, which further improves the overall performance of our method.

\section{Experiments}
\subsection{Datasets and Evaluation Metrics}
\noindent \textbf{ActivityNet Caption.}
Activity Caption \cite{krishna2017dense} contains 20000 videos with 100000 descriptions from YouTube \cite{caba2015activitynet}. The videos are 2 minutes on average, and these annotated video clips have much larger variation, ranging from several seconds to over 3 minutes. Since the test split is withheld for competition, following public split \cite{gao2017tall}, we use 37421, 17505, and 17031 sentence-video pairs for training, validation, and testing respectively.

\noindent \textbf{TACoS.}
TACoS is collected by \cite{regneri2013grounding} for video grounding and dense video captioning tasks. It consists of 127 videos on cooking activities with an average length of 4.79 minutes. In video grounding task, it contains 18818 video-query pairs. 
For fair comparisons, we follow the same split of the dataset as \cite{gao2017tall}, which has 10146, 4589, and 4083 video-query pairs for training, validation, and testing respectively.

\noindent \textbf{Charades-STA.}
Charades-STA is a benchmark dataset for the video grounding task, which is built upon the Charades \cite{sigurdsson2016hollywood} dataset. It is collected for video action recognition and video captioning, and contains 6672 videos and involves 16128 video-query pairs.
Following previous work \cite{gao2017tall}, we utilize 12408 video-query pairs for training and 3720 pairs for testing.

\noindent \textbf{Evaluation metric.}
We adopt “R@n, IoU=m” proposed by \cite{hu2016natural} as the evaluation metric, which calculates the IoU between the top-n retrieved video segments and the ground truth. It denotes the percentage of language queries having at least one segment whose IoU with ground truth is larger than m.
In our experiments, we use $n \in \{1,5\}$ for all datasets, $m \in \{0.5,0.7\}$ for ActivityNet Caption and Charades-STA, and $m \in \{0.3,0.5\}$ for TACoS.

\subsection{Implementation details}
Following \cite{zeng2020dense,liu2021context}, for video encoding, we apply a pre-trained C3D network \cite{tran2015learning} to obtain embedded features on all three datasets.
Besides, we also extract the I3D \cite{carreira2017quo} and VGG \cite{simonyan2014very} features on Charades-STA for fair comparison with some previous works. After that, we downsample the feature sequence of each video to 200 for ActivityNet Caption and TACoS, and 64 for Charades-STA. For sentence encoding, we utilize the Glove model \cite{pennington2014glove} to embed word-level features. The feature dimension $D$ is set to 512, and the head in multi-head self-attention is set to 8. The number of predicted positive frames is $K=5$.
The hyper-parameters $\alpha_1,\alpha_2$ in matching score are set to $0.6,0.4$ respectively, and the hyper-parameters $\gamma_1,\gamma_2$ in matching loss are set to $\gamma_1=1.0,\gamma_2=0.5$ respectively. The margin parameters are set to $\beta_1=\beta_2=0.2$. To infer the adjacent frames, we set the threshold as $\Theta=0.75$ in our all experiments.
During the training, we warm up our SL module with an Adam optimizer for 50 epochs, and train the BP module and finetune the whole SLP model for 50 and 100 epochs, respectively. The initial learning rate is set to 0.0001 and it is divided by 10 when the loss arrives on plateaus. 

\begin{table}[t!]
    \small
    \centering
    \caption{Performance comparison with the state-of-the-art NLVL models on the ActivityNet Caption dataset.}
    \setlength{\tabcolsep}{1.2mm}{
    \begin{tabular}{c|c|cccc}
    \toprule
    \multirow{2}*{Method} & \multirow{2}*{Feature} & R@1, & R@1, & R@5, & R@5, \\ 
    ~ & ~ & IoU=0.5 & IoU=0.7 & IoU=0.5 & IoU=0.7  \\ \midrule 
    CTRL \cite{gao2017tall} & C3D & 29.01 & 10.34 & 59.17 & 37.54  \\
    QSPN \cite{xu2019multilevel} & C3D & 33.26 & 13.43 & 62.39 & 40.78  \\
    SCDM \cite{yuan2019semantic} & C3D & 36.75 & 19.86 & 64.99 & 41.53  \\
    LGI \cite{mun2020LGI} & C3D & 41.51 & 23.07 & - & - \\
    VSLNet \cite{zhang2020span} & C3D & 43.22 & 26.16 & - & -  \\
    IVG-DCL \cite{nan2021interventional} & C3D & 43.84 & 27.10 & - & - \\
    CMIN \cite{zhang2019cross} & C3D & 43.40 & 23.88 & 67.95 & 50.73 \\
    DRN \cite{zeng2020dense} & C3D & 45.45 & 24.36 & 77.97 & 50.30 \\ 
    2DTAN \cite{zhang2019learning} & C3D & 44.51 & 26.54 & 77.13 & 61.96 \\
    SeqPAN \cite{zhang2021parallel} & C3D & 45.50 & 28.37 & - & - \\
    MATN \cite{zhang2021multi} & C3D & 48.02 & 31.78 & 78.02 & 63.18 \\
    CBLN \cite{liu2021context} & C3D & 48.12 & 27.60 & 79.32 & 63.41 \\
    GTR \cite{cao2021pursuit} & C3D & 50.57 & 29.11 & 80.43 & 65.14 \\
    \midrule
    \textbf{SLP} & C3D & \textbf{52.89} & \textbf{32.04} & \textbf{82.65} & \textbf{67.21} \\ \bottomrule
    \end{tabular}}
    \label{tab:sota1}
\end{table}

\subsection{Comparison with State-of-the-Arts}
\noindent \textbf{Compared methods.}
We compare the proposed SLP with state-of-the-art NLVL methods on three datasets. These methods are grouped into two categories by the viewpoints of proposal-based and proposal-free approach: (1) Proposal-based approach: CTRL \cite{gao2017tall}, QSPN \cite{xu2019multilevel}, SCDM \cite{yuan2019semantic}, CMIN \cite{zhang2019cross}, DRN \cite{zeng2020dense}, 2DTAN \cite{zhang2019learning}, CBLN \cite{liu2021context}, GTR \cite{cao2021pursuit}; 
These methods first sample multiple candidate video segments, and then directly compute the semantic similarity between the query representations with segment representations for ranking and selection. 
(2) Proposal-free approach: LGI \cite{mun2020LGI}, VSLNet \cite{zhang2020span}, IVG-DCL \cite{nan2021interventional}, SeqPAN \cite{zhang2021parallel}, MATN \cite{zhang2021multi}; 
These methods directly predict the start and end timestamps of the target segment by regression.
In all result tables, the scores of compared methods are reported in the corresponding papers.

\begin{table}[t!]
    \small
    \centering
    \caption{Performance comparison with the state-of-the-art NLVL models on the TACoS dataset.}
    \setlength{\tabcolsep}{1.2mm}{
    \begin{tabular}{c|c|cccc}
    \toprule
    \multirow{2}*{Method} & \multirow{2}*{Feature} & R@1, & R@1, & R@5, & R@5,  \\ 
    ~ & ~ & IoU=0.3 & IoU=0.5 & IoU=0.3 & IoU=0.5 \\ \midrule 
    CTRL \cite{gao2017tall} & C3D & 18.32 & 13.30 & 36.69 & 25.42 \\
    QSPN \cite{xu2019multilevel} & C3D & 20.15 & 15.23 & 36.72 & 25.30  \\
    SCDM \cite{yuan2019semantic} & C3D & 26.11 & 21.17 & 40.16 & 32.18 \\
    VSLNet \cite{zhang2020span} & C3D & 29.61 & 24.27 & - & - \\
    CMIN \cite{zhang2019cross} & C3D & 24.64 & 18.05 & 38.46 & 27.02 \\
    DRN \cite{zeng2020dense} & C3D & - & 23.17 & - & 33.36 \\ 
    SeqPAN \cite{zhang2021parallel} & C3D & 31.72 & 27.19 & - & - \\
    2DTAN \cite{zhang2019learning} & C3D & 37.29 & 25.32 & 57.81 & 45.04
    \\
    IVG-DCL \cite{nan2021interventional} & C3D & 38.84 & 29.07 & - & - \\
    CBLN \cite{liu2021context} & C3D & 38.98 & 27.65 & 59.96 & 46.24 \\
    GTR \cite{cao2021pursuit} & C3D & 40.39 & 30.22 & 61.94 & 47.73 \\
    \midrule
    \textbf{SLP} & C3D & \textbf{42.73} & \textbf{32.58} & \textbf{64.30} & \textbf{50.17} \\ \bottomrule
    \end{tabular}}
    \label{tab:sota2}
    \vspace{-10pt}
\end{table}

\begin{table}[t!]
    \small
    \centering
    \caption{Performance comparison with the state-of-the-art NLVL models on the Charades-STA dataset.}
    \setlength{\tabcolsep}{1.2mm}{
    \begin{tabular}{c|c|cccc}
    \toprule 
    \multirow{2}*{Method} & \multirow{2}*{Feature} & R@1, & R@1, & R@5, & R@5, \\ 
    ~ & ~ & IoU=0.5 & IoU=0.7 & IoU=0.5 & IoU=0.7 \\ \midrule 
    2DTAN \cite{zhang2019learning} & VGG & 39.81 & 23.25 & 79.33 & 51.15 \\
    DRN \cite{zeng2020dense} & VGG & 42.90 & 23.68 & 87.80 & 54.87 \\
    CBLN \cite{liu2021context} & VGG & 43.67 & 24.44 & 88.39 & 56.49 \\
    \textbf{SLP} & VGG & \textbf{45.83} & \textbf{26.15} & \textbf{90.52} & \textbf{58.31} \\ \midrule
    CTRL \cite{gao2017tall} & C3D & 23.63 & 8.89 & 58.92 & 29.57 \\
    QSPN \cite{xu2019multilevel} & C3D & 35.60 & 15.80 & 79.40 & 45.40 \\
    DRN \cite{zeng2020dense} & C3D & 45.40 & 26.40 & 88.01 & 55.38 \\
    CBLN \cite{liu2021context} & C3D & 47.94 & 28.22 & 88.20 & 57.47 \\
    \textbf{SLP} & C3D & \textbf{49.26} & \textbf{30.09} & \textbf{90.14} & \textbf{58.80} \\ \midrule
    DRN \cite{zeng2020dense} & I3D & 53.09 & 31.75 & 89.06 & 60.05 \\ 
    SCDM \cite{yuan2019semantic} & I3D & 54.44 & 33.43 & 74.43 & 58.08 \\
    LGI \cite{mun2020LGI} & I3D & 59.46 & 35.48 & - & - \\
    CBLN \cite{liu2021context} & I3D & 61.13 & 38.22 & 90.33 & 61.69 \\
    \textbf{SLP} & I3D & \textbf{64.35} & \textbf{40.43} & \textbf{92.68} & \textbf{63.22}  \\ \bottomrule
    \end{tabular}}
    \label{tab:sota3}
\end{table}

\noindent \textbf{Comparison on ActivityNet Caption}.
As shown in Table~\ref{tab:sota1}, we compare our SLP with the state-of-the-art proposal-based and proposal-free methods on ActivityNet Caption dataset, where we achieve a new state-of-the-art performance in terms of all metrics. Particularly, the proposed SLP model outperforms the best proposal-based method GTR
with 2.32\%, 2.93\%, 2.22\% and 2.07\% improvements on the all metrics, respectively. It also makes an
even larger improvement over the best proposal-free method MATN in metrics R@1, IoU=0.5 and R@1, IoU=0.5 by
4.87\% and 4.63\%. It verifies the benefits of utilizing two-step localization framework to predict the fine-grained positive frames and capture the detailed differences among the consecutive frames.

\noindent \textbf{Comparison on TACoS}.
We also compare SLP with the state-of-the-art proposal-based and proposal-free methods on the TACoS dataset in Table~\ref{tab:sota2}.
On TACoS dataset, the cooking activities take place in the same kitchen scene with slightly varied cooking ob- jects, thus showing the challenging nature of this dataset. Despite its difficulty, we still reach the highest results over all evaluation metrics. Particularly, our SLP outperforms the best proposal-based method GTR by 2.36\% and 2.44\% absolute improvement in terms of R@1, IoU=0.7 and R@5, IoU=0.7, respectively. Compared to the proposal-free method IVG-DCL, it outperforms it by 3.89\% and 3.51\% in terms of R@1, IoU=0.5 and R@1, IoU=0.7, respectively.

\noindent \textbf{Comparison on Charades-STA}.
Table~\ref{tab:sota3} also report the comparison of localization results on Charades-STA dataset.
For fair comparison with different methods, we perform experiments with same features (\ie, VGG, C3D, and I3D) reported in their papers. We can find that our SLP reaches the highest results over all evaluation metrics.
Specifically, when using the same VGG features, compared to the previously best method CBLN, our model brings the absolute improvement of 2.16\%, 1.71\%, 2.13\% and 1.82\% on all metrics, respectively. 
When using the same C3D or I3D features, it is obvious that our model still performs better than the existing methods.
All these results again verify the effectiveness of our model.

\subsection{Ablation Study}
In this section, we perform in-depth ablation studies to analyze the effectiveness of our proposed SLP. Specifically, We first conduct the main ablation study to investigate the contribution of each components, then examine the effectiveness of SL and BP modules, and analyze the impact of different settings on hyper-parameters.

\begin{table}[t!]
    \small
    \centering
    \caption{Main ablation studies on the ActivityNet Caption dataset, where `MME" and `QCC" denote the multi-modal encoders and query-guided content comprehension.}
    \setlength{\tabcolsep}{2.0mm}{
    \begin{tabular}{c|ccc|cc}
    \toprule 
    \multirow{2}*{Model} & \multicolumn{2}{c}{SL Module} & \multirow{2}*{BP Module} & R@1, & R@1, \\ \cline{2-3}
    ~ & MME & QCC & ~ & IoU=0.5 & IoU=0.7 \\ \midrule
    \ding{172} & $\checkmark$ & $\times$ & $\times$ & 38.95 & 19.62 \\ \midrule
    \ding{173} & $\checkmark$ & $\checkmark$ & $\times$ & 42.78 & 23.40 \\
    \ding{174} & $\checkmark$ & $\times$ & $\checkmark$ & 47.31 & 27.59\\
    \ding{175} & $\checkmark$ & $\checkmark$ & $\checkmark$ & \textbf{52.89} & \textbf{32.04}
    \\ \bottomrule
    \end{tabular}}
    \label{tab:ablation1}
    \vspace{-10pt}
\end{table}

\begin{table}[t!]
    \small
    \centering
    \caption{Investigation on the Skimming-and-Locating (SL) module on the ActivityNet Caption dataset.}
    \setlength{\tabcolsep}{2.0mm}{
    \begin{tabular}{c|c|cc}
    \toprule 
    \multirow{2}*{Components} & \multirow{2}*{Variants} & R@1, & R@1, \\
    ~ &  ~ & IoU=0.5 & IoU=0.7 \\ \midrule
    \multirow{2}*{MME} & w/o transformer & 50.27 & 30.53 \\
    ~ & w/ transformer & \textbf{52.89} & \textbf{32.04} \\ \midrule
    \multirow{3}*{QCC} & graph layer=0 & 48.07 & 28.82\\
    ~ & graph layer=1 & \textbf{52.89} & \textbf{32.04} \\
    ~ & graph layer=2 &  51.16 & 30.95
    \\ \bottomrule
    \end{tabular}}
    \label{tab:ablation2}
\end{table}

\noindent \textbf{Main ablation study.}
As shown in Table~\ref{tab:ablation1}, we verify the contribution of each component in our SLP. For SL module, it consists of a multi-modal encoder (MME) and the query-guided content comprehension module (QCC). We first implement the baseline model \ding{172} by directly regressing the start/end timestamps based on the encoded video and query features. In this model, we simply utilize a common co-attention mechanism \cite{lu2016hierarchical} for multi-modal interaction and adopt the boundary regression head as in \cite{zhang2020span}.
After adding the QCC module in model \ding{173}, we find that QCC effectively enhances the discriminative frame-wise representation learning, bringing a large improvement gains of 3.83\% and 3.78\% in terms of R@1, IoU=0.5 and R@1, IoU=0.7. 
Here, the model \ding{173} can be taken as a variant of a single-step model.
By combing the Sl module \ding{173} with the BP module to model \ding{175}, there are the most significant improvement of 10.11\% and 8.64\% since the BP module captures more fine-grained differences between consecutive frames than the single-step regression head. It demonstrates that our two-step framework is effective to generate more accurate target segments.

\noindent \textbf{Ablation study on the SL module.}
As shown in Table~\ref{tab:ablation2}, we first investigate the impact on different variants of multi-modal encoders (MME). As the multi-head self-attention in MME is a variant of transformer encoder, we remove it in both video and query encoders for comparison. It shows that w/ transformer brings significant improvement since it captures the long-range dependencies among frames/words. Then, we investigate the impact on different layers of the multi-modal graph in QCC. As shown in Table~\ref{tab:ablation2}, comparing with the model w/o multi-modal graph (graph layer=0), our full model w/multi-modal graph is effective to enhance the discriminative frame-wise feature learning for better classifying positive-negative frame, and achieves the best result with a single layer. More graph layers will result in over-smoothing \cite{li2018deeper} problem.

\begin{figure*}[t!]
\centering
\includegraphics[width=1.0\textwidth]{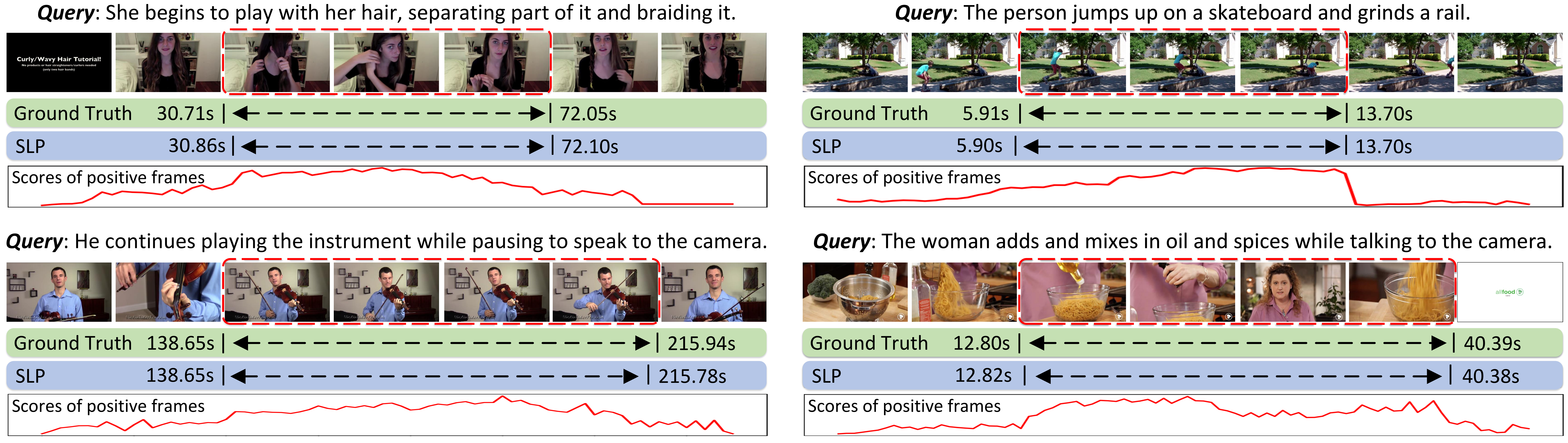}
\vspace{-10pt}
\caption{Qualitative results of our SLP on the ActivityNet Caption dataset. The red curve denotes the frame-wise score obtained by the binary classification function in SL module for predicting the positive frames.}
\vspace{-10pt}
\label{fig:result}
\end{figure*}

\begin{table}[t!]
    \small
    \centering
    \caption{Investigation on the Bi-directional Perusing (BP) module on the ActivityNet Caption dataset.}
    \setlength{\tabcolsep}{2.0mm}{
    \begin{tabular}{c|c|cc}
    \toprule 
    \multirow{2}*{Components} & \multirow{2}*{Variants} & R@1, & R@1, \\
    ~ &  ~ & IoU=0.5 & IoU=0.7 \\ \midrule
    \multirow{3}*{\tabincell{c}{Directional\\Perusing}} & left-then-right & \textbf{53.06} & 31.81\\
    ~ & right-then-left & 52.93 & 31.97 \\
    ~ & left-while-right & 52.89 & \textbf{32.04} \\ \midrule
    \multirow{3}*{\tabincell{c}{Segment\\Update}} & w/ max-pooling & 49.51 & 28.20 \\
    ~ & w/ concatenation & 51.66 & 30.75 \\
    ~ & w/ update & \textbf{52.89} & \textbf{32.04}
    \\ \bottomrule
    \end{tabular}}
    \vspace{-10pt}
    \label{tab:ablation3}
\end{table}

\noindent \textbf{Ablation study on the BP module.}
For BP module, we first investigate the impact of different directions of the perusing process: 1) first perusing the left frames until no frame matches the target semantic (``left-then-right"); 2) first perusing the right frames until no frame matches the target semantic (``right-then-left"); 3) perusing the left and right frames at the same time (``left-while-right"). As shown in Table~\ref{tab:ablation3}, there is no significant difference in the performances of different perusing directions, which indicates our SLP model is insensitive to the directions of perusing. Therefore, we utilize the ``left-while-right" perusing direction in all experiments. 
We also investigate the effectiveness of different strategies for integrating features of new input frame with the current segment. It shows that our learnable selective segment updating strategy performs better than both max-pooling and concatenation operations.

\begin{table}[t!]
    \small
    \centering
    \caption{Sensitivity analysis on the hyper-parameters $K$, $\alpha_1,\alpha_2$ and $\Theta$ on the ActivityNet Caption dataset.}
    \setlength{\tabcolsep}{2.0mm}{
    \begin{tabular}{c|cccc|cc}
    \toprule 
    \multirow{2}*{Variable} & \multirow{2}*{$K$} & \multirow{2}*{$\alpha_1$} & \multirow{2}*{$\alpha_2$} & \multirow{2}*{$\Theta$} & R@1, & R@1, \\
    ~ & ~ & ~ & ~ &  ~ & IoU=0.5 & IoU=0.7 \\ \midrule
    \multirow{4}*{$K$} & 1 & 0.6 & 0.4 & 0.75 & 49.85 & 30.14 \\
    ~ & 3 & 0.6 & 0.4 & 0.75 & 51.57 & 31.23 \\
    ~ & 5 & 0.6 & 0.4 & 0.75 & 52.89 & \textbf{32.04} \\
    ~ & 7 & 0.6 & 0.4 & 0.75 & \textbf{52.92} & \textbf{32.04} \\ \midrule
    \multirow{6}*{$\alpha_1,\alpha_2$} & 5 & 0.0 & 1.0 & 0.75 & 47.69 & 27.17 \\
    ~ & 5 & 0.2 & 0.8 & 0.75 & 50.06 & 29.40 \\
    ~ & 5 & 0.4 & 0.6 & 0.75 & 51.77 & 31.23 \\
    ~ & 5 & 0.6 & 0.4 & 0.75 & \textbf{52.89} & \textbf{32.04} \\
    ~ & 5 & 0.8 & 0.2 & 0.75 & 52.35 & 31.62 \\
    ~ & 5 & 1.0 & 0.0 & 0.75 & 51.48 & 30.91 \\ \midrule
    \multirow{6}*{$\Theta$} & 5 & 0.6 & 0.4 & 0.60 & 48.59 & 29.02 \\
    ~ & 5 & 0.6 & 0.4 & 0.65 & 50.17 & 30.23 \\
    ~ & 5 & 0.6 & 0.4 & 0.70 & 51.44 & 31.35 \\
    ~ & 5 & 0.6 & 0.4 & 0.75 & \textbf{52.89} & \textbf{32.04} \\
    ~ & 5 & 0.6 & 0.4 & 0.80 & 51.60 & 31.47 \\
    ~ & 5 & 0.6 & 0.4 & 0.85 & 51.01 & 30.58
    \\ \bottomrule
    \end{tabular}}
    \vspace{-10pt}
    \label{tab:ablation4}
\end{table}

\noindent \textbf{Impact of hyper-parameters.}
We first analyze the impact of different number $K$ of the predicted top ranked positive frames. As shown in Table~\ref{tab:ablation4}, a larger $K$ results in more accurate localization, leading to precise segment boundary. This is because some false-positive predicted frames degenerate the performance when the number $K$ is small. The model with $K=7$ achieves the best result but only performs marginally better than that with $K=5$ at the expense of significantly larger cost of GPU memory and time.
Therefore, we choose $K=5$ in all the experiments. 
We then analyze the balancing weights $\alpha_1,\alpha_2$ of the linguistic and visual similarity scores in Equation (\ref{eq:score}). Table~\ref{tab:ablation4} shows that both scores are crucial to the final matching, where the model with $\alpha_1=0.6,\alpha_2=0.4$ achieves the best result. It also shows that the linguistic score is a bit more important than the visual one, since some complicated visual appearances among consecutive frames are harder to distinguish.
At last, we evaluate the model with different thresholds $\Theta$ to investigate its impact on adjacent frame definition (should add into the current segment or not). Specifically, a larger $\Theta$ denotes a stricter segment construction. From the table, we see that the model with $\Theta=0.75$ performs the best. Overall, the performance changes only a small amount as the hyperparameters are changed.

\subsection{Visualization Results}
To qualitatively validate the effectiveness of our SLP, we investigate the localization results of several typical examples as shown in Figure~\ref{fig:result}. By deploying the ``Skimming-Locating-Perusing" human-like localization strategy, our SLP model captures more fine-grained differences between adjacent frames, thus leading to produce precise segment boundaries. Besides, to investigate the positive-negative classification performance of the SL module, we also plot the scores of the video frames predicted by Equation (\ref{eq:loss1}) in Figure~\ref{fig:result} (red curve). We can find that the frames with highest scores are almost fall into the ground-truth segment, which demonstrates that the SL module is able to locate the positive frames well.

\section{Conclusion}
In this paper, we argued that existing NLVL methods follow the unnatural ``only look once" framework and overlook two indispensable characteristics: frame-differentiable and boundary-precise. 
To this end, we propose a human-like framework called Skimming-Locating-Perusing (SLP), which consists of a Skimming-and-Locating (SL) module and a Bi-directional Perusing (BP) module. The SL module first comprehends the query semantic and interacts it with the video features to distinguish positive and negative frames. Then, the BP module initializes the segment based on each positive frame and dynamically updates it by adding adjacent frames sharing the same semantic. 
With such "Skimming-Locating-Perusing" localization strategy, we are able to determine more precise segment boundaries.
Experimental results on three challenging benchmarks demonstrate the superiority of our SLP compared to the state-of-the-arts.
\bibliographystyle{ACM-Reference-Format}
\bibliography{sample-base}


\end{document}